# Correcting Errors in Digital Lexicographic Resources Using a Dictionary Manipulation Language

**David Zajic*†, Michael Maxwell†, David Doermann*, Paul Rodrigues†, Michael Bloodgood†**

†University of Maryland Center for Advanced Study of Language (CASL)
*University of Maryland Institute for Advanced Computer Studies (UMIACS)
College Park, MD
E-mail: dzajic@casl.umd.edu, mmaxwell@casl.umd.edu, doermann@umiacs.umd.edu, prr@umd.edu, meb@umd.edu

**Abstract**

We describe a paradigm for combining manual and automatic error correction of noisy structured lexicographic data. Modifications to the structure and underlying text of the lexicographic data are expressed in a simple, interpreted programming language. Dictionary Manipulation Language (DML) commands identify nodes by unique identifiers, and manipulations are performed using simple commands such as *create*, *move*, *set text*, etc. Corrected lexicons are produced by applying sequences of DML commands to the source version of the lexicon. DML commands can be written manually to repair one-off errors or generated automatically to correct recurring problems. We discuss advantages of the paradigm for the task of editing digital bilingual dictionaries.

## 1. Introduction

Digital lexicographic resources are frequently derived from print dictionaries, either manually or automatically, or are adapted from publishers' files. Often the resulting digital lexicographic resources contain errors. Discovering and correcting errors in lexicographic data is a common task for teams dealing with digital lexicographic resources. We propose a paradigm for correcting errors and discuss its advantages for the task of editing digital bilingual dictionaries.

A digitized dictionary contains not only the underlying text, but also structural information. The underlying text is divided into meaningful spans and the spans are organized into a structure that denotes the relationships among them. Print dictionaries denote structural information with fonts, indentation, special symbols, and other visual clues. In a digitized version of a print lexicon, the structural information is made explicit. One goal for editing structured lexicographic data is to ensure that the structural information matches the semantics implicit in the layout of the source print lexicon.

The process of repairing a digital dictionary includes correcting text errors, such as typos and OCR errors, and structural errors. Structural errors happen when the underlying text is split into text spans incorrectly or when the relationships among text spans are incorrect. We refer to data containing these types of errors as noisy structured data, because our goal is to recover the true representation of the dictionary contents by correcting errors introduced by the noisy process of digitizing it.

For example, in Qureshi (1971), an Urdu to English dictionary, the translation of Urdu word "بیجو" is "goal in children's game called باؤری." In the source digitization this translation was split into "goal in children's game called" and a separate lexical entry "باؤری." The underlying text was split incorrectly into two distinct text spans.

In some cases the underlying text is divided into correct spans, but the role of the text spans is incorrect. The translation of Urdu "شاذ و نادر" is "rarely," but in the source digitization, "rarely" was identified as a usage note rather than a translation.

There are also cases in which the text is divided and tagged with the correct role, but its relationship to other text spans is incorrect. For example, in Qureshi (1971) lexical entries are organized into blocks of text containing a headword followed by collocations containing the headword. We observed that in the digitized version of the paragraph for "ٹانگ", "leg", the translation of the phrase "ٹانگ تلے سے نکلنا", "to yield, submit" was attached to the headword instead of to the phrase.

Another goal of editing lexical resources is to map resource- and language-specific structures into resource- and language-independent standards, such as Text Encoding Initiative (TEI) (Ide & Véronis, 1995) or Lexical Markup Framework (LMF) (Francopoulo et al., 2006).

## 2. Database and Version Control Solutions

A straightforward method of editing a structured digital resource is to store the information in a shared repository and allow experts to edit the repository contents. The repository could be a relational database or an XML document under a version control system. When the resource is in a database users modify the data through a transaction processing system. For a document under version control, users check out a working copy of the resource, edit the copy, and commit their edited copies to the repository.

```
CREATE TextElement tag text relation anchor
CREATE Element tag relation anchor
CREATE Clone source relation anchor
REMOVE Element target
REMOVE Text target
RETAG target tag
MOVE Element target relation anchor
SET Attribute target attribute value
SET Text target text
```

Figure 1: A partial list of DML commands.

Suppose a team wishes to undo a local change to the resource. In both approaches it is straightforward to restore the resource to its state at a specific time, but all changes made subsequent to that time are lost. We propose a paradigm that supports non-chronological rollback of local operations, which would allow for undo of a local edit while preserving subsequent human effort.

At some point during or after the lexicon repair process a team might wish to analyze the changes made to the resource. The transaction information stored by database or version control approaches would require substantial processing to clearly represent the changes from the initial to the final versions of the resource. Our paradigm creates an executable record of the modifications necessary to convert the source resource into the final version.

If the editing process is a *lossy* transformation of the source into a standard format, meaning that it is not possible to reconstruct the original data from the transformed data, it is desirable to preserve a copy of the original source. Our paradigm makes preservation of the original source an integral part of the editing process.

## 3. Dictionary Manipulation Language (DML) Paradigm

The key intuition of our paradigm for editing digital lexicographic resources is that the edits take the form of commands in DML rather than direct modifications to a shared resource. DML commands can be written manually by language experts or generated automatically by computer systems. The end-to-end process of generating a final lexical resource from the original source consists of reading the original source lexicon into computer memory, applying a sequence of DML command sets to it, and writing the result to a destination resource. The original source file is never edited directly. Instead the DML command sets are edited by language experts for unique problems based on examination of the source lexicon, or an interim state of the lexicon. DML command sets are also generated at run-time to correct repeated problems and then applied to the in-memory resource.

DML commands are applied to a lexicon by a DML interpreter program. The interpreter loads an XML

```
ENTRY ID="351782">
  <FORM ID="351783">
    <ORTH ID="351784">طرفہ</ORTH>
    <PRON ID="351785">tūr'fah</PRON>
  </FORM>
  ...
  <SENSE N="3" ID="351794">
    <USG TYPE="time" ID="351795">rare</USG>
  </SENSE>
  ...
</ENTRY>

# ABC 5/27/2011 sense tagged as usage, retag
CREATE  element   TRANS   under   351794   T
RETAG   351795   TR
REMOVEattribute   351795   TIME
MOVE    element   351795   under   T

  <ENTRY ID="351782">
    <FORM ID="351783">
      <ORTH ID="351784">طرفہ</ORTH>
      <PRON ID="351785">tūr'fah</PRON>
    </FORM>
    ...
    <SENSE N="3" ID="351794">
      <TRANS ID="351794+1">
        <TR ID="351795">rare</TR>
      </TRANS>
    </SENSE>
    ...
  </ENTRY>
```

Figure 2: XML excerpt from Urdu dictionary, DML commands to correct a structural error, and the result of applying the DML commands to the source XML.

lexicon into memory, reads the DML commands from one or more DML files, performs the operations denoted by the DML commands on the in-memory lexicon, and writes the result to an XML file.

The work of lexicon repair consists of writing the DML commands to correct unique problems and writing programs that search for repeated problems and generate DML commands to correct them.

### 3.1 DML Commands

DML commands operate on XML documents in which each element of the XML document has a unique identifier. Figure 1 shows a partial list of DML commands.

Figure 2 shows an excerpt from an XML document containing a structural error, some DML commands to correct the error, and the result of applying the DML commands to the source. In this instance, annotator ABC observed that the translation of the third sense of "طرفہ" should be "rare," instead of the usage note that the third sense is a rare meaning. She wrote a comment about her observation and her intended solution, then solved the problem by creating a new TRANS element, changing the element tag of USG to TR, moving the TR inside the new TRANS element, and removing the TIME attribute from the TR element.

### 3.2 DML Processing

The end-to-end process by which a source lexicon is converted to a final lexicon consists of reading the source lexicon from an XML file into memory, applying a sequence of DML command sets to it and writing the output to an XML file. As a diagnostic option, the interim state of the lexicon can be written to an XML file after the application of any or all of the DML command sets. The architecture is shown in Figure 3.

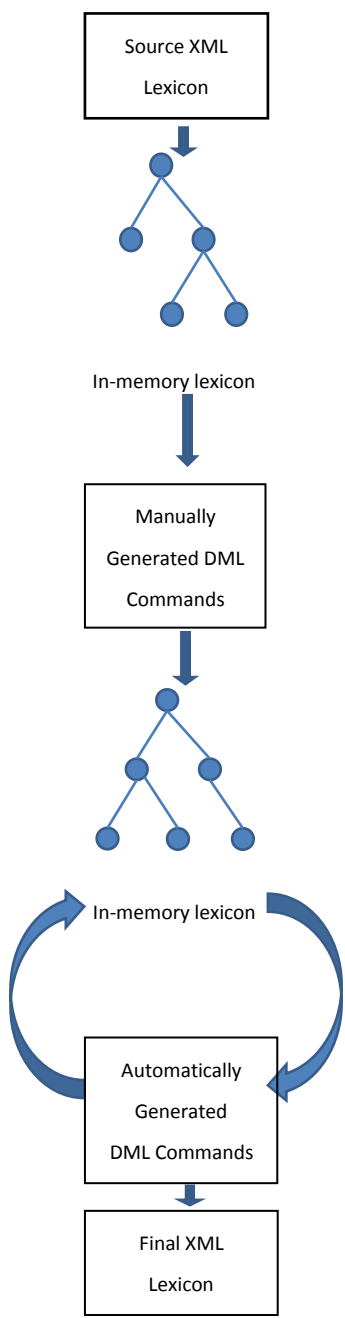

Figure 3 : Architecture of DML processing.

A DML command set is a file containing DML commands that solve a collection of similar problems. A DML command set could convert the text contents of certain tags from non-standard legacy encodings to Unicode, or could relocate pieces of grammatical information that repeatedly and consistently appear in the wrong position.

Some DML command sets are written by language experts to correct unique problems while other DML command sets are generated at run time by applying patterns to the current state of the lexicon, and are then applied to the lexicon. For example, a module can search for all instances in which a word sense element has been incorrectly split from its lexical entry. Every time it finds one, the module generates DML commands to move the sense into its proper place. When the DML command set to solve this sort of problem is complete, the DML command set is applied to the in-memory lexicon. Thus we have a lexicon with the problem corrected, and we have a record of the specific changes that corrected the instances of the problem.

## 4. Advantages of the DML paradigm

This section will describe the advantages that motivate the use of DML for editing noisy structured lexicons.

### 4.1 Preservation of Source Data

The motivation for correcting errors in a lexicon is frequently to prepare the lexicon for use in a specific task. For example, a lexicon repair team might wish to load a dictionary's contents into an enterprise-wide dictionary interface that requires a specific data format. A work paradigm that converts the source into the desired format by directly altering the source can cause the loss of valuable information. If it is later discovered that a significant error was made in generating the target lexicon, it is critical to have access the original source. Because the DML paradigm is based on application of DML commands to the source lexicon, it encourages preservation of the original source data and discourages direct editing of the source data.

### 4.2 Non-chronological rollback

Under the DML paradigm it is possible to undo a local change to the lexicon without affecting changes that were performed afterwards. Using a database or a version control system, it is possible to restore a resource to its state at a particular time. A change is undone by restoring the resource to a time before the change was made.

Under the DML paradigm, the source lexicon is never directly edited, so a change is undone by removing the relevant DML commands and rerunning the process to generate the final lexicon.

### 4.3 Effect of Application Order for DML Command Sets

DML command sets to solve repeated problems are generated at run time and are then applied to the in-memory lexicon. It is possible for application of a manually written DML command set to change some part of the lexicon so that it matches the pattern that triggers automatic DML generation. The automatically generated DML commands are not just created once and applied in that form in subsequent runs. They are recreated with every run, so they can take advantage of repairs to the lexicon made by earlier DML command sets in searching for their trigger patterns. In a similar manner, a repair to the lexicon can prevent the inappropriate application of automatically generated DML by changing a section of the lexicon so it does not match a trigger pattern. The effect of application order on DML command sets is similar to the ideas of feeding order and bleeding order of phonological rules.

The effect of ordering DML command sets can save effort for language experts. Finding a pattern in a lexicon can trigger the automatic generation of a large and complex group of DML commands. If a language expert can make a small local change in a lexicon so that so that it correctly matches the trigger pattern of a DML command generator, it is less work than manually writing the DML commands for the complex repair

### 4.4 DML as Documentation

The DML command sets serve as documentation of the changes that were made to the lexical resource. After an end-to-end run of a lexicon repair process, the manually and automatically generated DML command set files and the interim snapshots of the lexicon as XML files after the application of each DML command set remain as evidence of the changes that were made to the lexicon. One can examine all the changes to a local region of the lexicon by searching for element identifiers in the DML command files. Alternatively one can examine the effect of a DML command set by comparing the interim XML snapshots of the lexicon before and after the application of that set.

### 4.5 DML as Data

The DML command sets themselves can be analysed to better understand the process of discovering and correcting structural errors in digital lexicons. The DML command sets serve as training and evaluation data for research on machine learning systems. Our group is currently developing systems to automatically locate structural anomalies in digital lexicons (Rodrigues et al., 2011).

### 4.6 Support for Collaboration between Language Experts and Computer Scientists

We have found that language experts with no previous experience with computer scripting or XML data were able to learn how to write DML commands to make repairs to lexicons. This improved the workflow for finding and correcting structural errors since the same language expert who discovered an error could immediately correct it. This removed the bottleneck of creating a queue of errors and corrections to be implemented later by computer scientists. It also allowed the communication between the language experts and computer scientists on the lexicon repair team to focus on discovering and automatically correcting repeated error patterns.

### 5. Applications of DML

CASL's lexicon repair team has used the DML paradigm to perform structural repair on digital sources for three bilingual dictionaries: Iraqi Arabic to English (Woodhead & Beene 2003), Yemeni Arabic to English (Qafisheh 2000) and Urdu to English (Qureshi 1971). Table 1 shows the rough scope of these projects. These projects included restructuring the data to be compatible with LMF's resource- and language-independent schema for bilingual lexicons, and conversion of non-Latin text from legacy encodings to Unicode. The number of manual commands gives an idea of the scope of the human effort to correct unique textual and structural errors.

| Lexicon | Entries | DML commands |
|---------|---------|--------------|
| Iraqi | 13,719 | Manual: 4759<br>Automatic: 1,594,688 |
| Yemeni | 16,069 | Manual: 16,069<br>Automatic: 162,685 |
| Urdu | 44,237 | Manual: 5,963<br>Automatic: 707,612 |

Table 1: Numbers of lexical entries, manually written DML commands and automatically generated DML commands for three lexicon repair projects.

### 6. Future Work

We have found that DML is easy to use by language experts, however it does require the overhead of using a programmer's editor and directly examining XML documents. We are developing a graphical interface that will allow language experts to view the lexicon as a tree and perform correction operations through the interface, which would generate the DML commands to implement the changes. The interface would also allow users to see the effect on the tree of applying specific DML commands and to view similar areas of the lexicon to determine if similar corrections should be applied.

### 7. Conclusion

We have described a paradigm for editing noisy structured lexicographic data using DML. This approach addresses the problems of non-chronological rollback and preservation of original source data. It also offers the advantages that the DML command sets serve as documentation of the corrections made to the lexicon, and be used as training and testing data in research in automatic detection of anomalies in structured lexicographic data.

## 8. Acknowledgements

The authors wish to thank the language experts who served as the testers and users of DML: Bridget Hirsch, Christian Hettick, Mohini Madgavkar, Valerie Novak, Rebecca McGowan, and Jennifer Boutz.

This material is based upon work supported, in whole or in part, with funding from the United States Government. Any opinions, findings and conclusions, or recommendations expressed in this material are those of the author(s) and do not necessarily reflect the views of the University of Maryland, College Park and/or any agency or entity of the United States Government. Nothing in this report is intended to be and shall not be treated or construed as an endorsement or recommendation by the University of Maryland, United States Government, or the authors of the product, process, or service that is the subject of this report. No one may use any information contained or based on this report in advertisements or promotional materials related to any company product, process, or service or in support of other commercial purposes.